# STABILIZATION CONTROL OF THE DIFFERENTIAL MOBILE ROBOT USING LYAPUNOV FUNCTION AND EXTENDED KALMAN FILTER

**Thuan Hoang Tran, Manh Duong Phung, Thi Thanh Van Nguyen ,
Quang Vinh Tran**

*University of Engineering and Technology, VNU, Hanoi City, Vietnam*

Corresponding author: *thuanhoang@donga.edu.vn*

## ABSTRACT

This paper presents the design of a control model to navigate the differential mobile robot to reach the desired destination from an arbitrary initial pose. The designed model is divided into two stages: the state estimation and the stabilization control. In the state estimation, an extended Kalman filter is employed to optimally combine the information from the system dynamics and measurements. Two Lyapunov functions are constructed that allow a hybrid feedback control law to execute the robot movements. The asymptotical stability and robustness of the closed loop system are assured. Simulations and experiments are carried out to validate the effectiveness and applicability of the proposed approach.

## 1. INTRODUCTION

Reliable navigation is the key problem in autonomous mobile robotics and it can be split into two categories corresponding to indoor and outdoor environments [1-6]. Success in navigation requires success at the four building blocks of navigation: perception, the robot must interpret its sensors to extract meaningful data; localization, the robot must determine its position in the environment; cognition, the robot must decide how to act to achieve its goals; and motion control, the robot must modulate its motor outputs to achieve the desired trajectory [7].

Of these four components, motion control has received great research attention due to the challenge in robot model. In general, the dynamics and kinematics of the mobile robot are nonlinear and nonholonomic (a system whose state depends on the path taken to achieve it) and consist of uncertainty parameters. The design of controller therefore requires nonlinear and statistical approaches. A number of methods to stabilize a nonholonomic system via feedback control have been proposed in the literature [8-13]. Most paper, however, assumed ideal condition in which there is no disturbances on the mobile robot system. In [13], feedback laws that globally exponentially stabilize the mobile robot with no input disturbances and measurement noises is proposed. Globally stabilizing time-varying feedbacks for nonholonomic systems are derived in [8] in which chain form systems were introduced to model the kinematics of the mobile robot. A discussion of their convergence properties in ideal condition is provided.

In practice, the noises arisen from the system kinematics and measurement devices are unavoidable so the robustness issue against actuator disturbances and measurement noises deserves further attention. Several methods have been proposed to study the robust stabilization of the nonholonomic system [14, 16, 17, 18-24]. In [17], an adaptive sliding-mode dynamic controller for wheeled mobile robots is designed and implemented. The PI-type sliding surface is employed for the first order nonlinear function of dynamics model and the convergence of the complete equations of motion is proven by the Lyapunov stability theory. In [14-16], the navigation variables transformed from configuration variables are introduced. They are the distance from the robot frame to the target frame, the angle between the robot-to-target vector and the target frame, and the angle between the robot-to-target vector and the current vehicle orientation. With these definitions, a stabilization control method that provides a fast and natural performance path is conducted.

In this paper, the problem of motion control for wheeled mobile robots is addressed. Starting from an initial configuration (position and orientation), the goal of the controller is to navigate the robot to reach a pre-defined final configuration in a manner that both the robot dynamics and measurements are subjected to uncertainty noises. A hybrid approach with the combination of statistical state estimation and Lyapunov stabilization control is proposed. First, the kinematics of the robot is reviewed and the configuration variables of the robot are reformulated in the form of navigation variables. A closed-loop feedback control is then constructed in which the Lyapunov function is employed to derive the control law while producing an optimal trajectory comprised of segments and circular arcs across objective points. Since the efficiency of the controller under noise condition mainly depends on the system feedback, an extended Kalman filter (EKF) is utilized to optimally estimate the system state by combining the measurement data and system dynamics [26-27].

The main contribution of the paper are (i) a stabilization controller is proposed for the differential mobile robot system with external disturbances; (ii) the stability analysis for the complete equations of the motion of the mobile robot are proven by the Lyapunov stability theory; (iv) Simulation and experiment results are used to illustrate the effectiveness of the proposed control schemes.

The paper is arranged as follows. Details of the control problem are described in Section II. The algorithm for state estimation using EKF is explained in Section III. Section IV introduces the design and implementation of the stabilization controller. Simulations and experiments are presented in section V. The paper concludes with an evaluation of the system, with respect to its strengths and weaknesses, and with suggestions of possible future developments.

## 2. PROBLEM FORMULATION

Consider the scenario shown in fig.1A, with an arbitrary position and orientation of the robot and a predefined goal position and orientation. The actual pose error vector between the initial and the final configuration given in the robot reference frame $\{X_R, Y_R, \theta\}$ is e=$^R[x, y, \theta]^T$ with x, y and θ being the goal coordinates of the robot.

The task of the controller layout is to find a control constraint, if it exists, of the translational and the angular velocity such that the error e is driven toward zero $\lim_{t \to \infty} e(t) = 0$

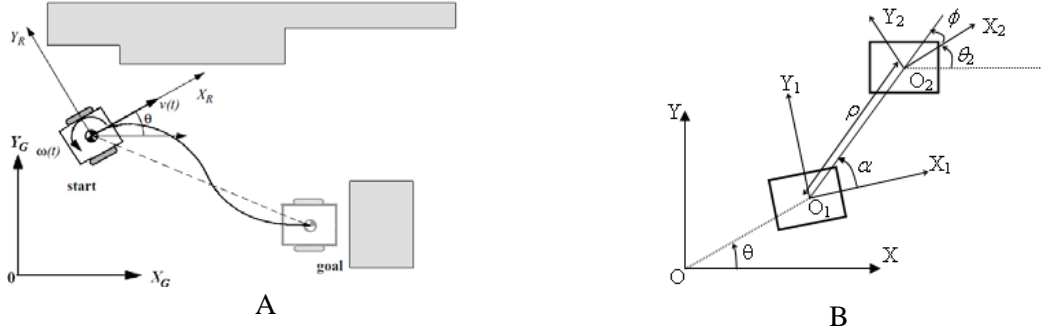

*Figure 1.* (A) The goal of the controller; (B) The robot poses and parameters

To formulate the problem in more details, we consider the two wheeled, differential-drive mobile robot with non-slipping and pure rolling. The kinematics of the described robot is given as (1).

$$\begin{cases} \dot{X} = v\cos\theta \\ \dot{Y} = v\sin\theta \\ \dot{\theta} = \omega \end{cases} \quad (1) \qquad \begin{cases} \dot{\rho} = -v\cos\alpha \\ \dot{\alpha} = -\omega + v\dfrac{\sin\alpha}{\rho} \\ \dot{\phi} = v\dfrac{\sin\alpha}{\rho} \end{cases} \quad (2)$$

where $\omega$ and $v$ are the control inputs which are respectively the rotational angular and the translational speed of the robot; $X, Y, \theta$ are the coordinates and the orientation of the robot in the global coordinate frame $OXY$.

As shown in fig.1B, Let $\rho$ be the distance between $O_1$ and $O_2$, $\phi$ be the angular made by the vector connecting $O_1$ and $O_2$ and the vector connection $O_2$ and $X_2$, $\alpha$ be the angular made by the vector connection $O_1$ and $O_2$ and the vector connection $O_1$ and $X_1$:

$$\begin{aligned} \rho &= \sqrt{(X_2 - X)^2 + (Y_2 - Y)^2} \\ \phi &= \operatorname{atan2}(Y_2 - Y, X_2 - X) - \theta_2 \\ \alpha &= \operatorname{atan2}(Y_2 - Y, X_2 - X) - \theta \end{aligned} \quad (3)$$

The kinematic equations in the navigation variables domain $(\rho, \alpha, \phi)$ are written as (2).

Let $(\hat{X}, \hat{Y}, \hat{\theta})$ and $(X, Y, \theta)$ be respectively the estimate and the real poses of the robot. Let $(\varepsilon_X, \varepsilon_Y, \varepsilon_\theta)$ be the estimate noises of the robot pose $(X, Y, \theta)$. The estimate values of the position $(\hat{X}, \hat{Y})$ and orientation $\hat{\theta}$ are defined as follows: $\hat{X} = X + \varepsilon_X$, $\hat{Y} = Y + \varepsilon_Y$, $\hat{\theta} = \theta + \varepsilon_\theta$ where $|\varepsilon_X| \leq \|\varepsilon_X\|$, $|\varepsilon_Y| \leq \|\varepsilon_Y\|$, $|\varepsilon_\theta| \leq \|\varepsilon_\theta\|$ are the absolute maximum values of the measurement noises of the position $(\hat{X}, \hat{Y})$ and orientation $\hat{\theta}$, respectively. Let $\varepsilon_\rho, \varepsilon_\alpha, \varepsilon_\phi$ denote the state feedback disturbances of the navigation variables $(\rho, \alpha, \phi)$:

$$\varepsilon_\rho = \sqrt{(X_2-\hat{X})^2+(Y_2-\hat{Y})^2} - \sqrt{(X_2-X)^2+(Y_2-Y)^2}$$

$$\varepsilon_\phi = \sqrt{(X_2-\hat{X})^2+(Y_2-\hat{Y})^2} - \sqrt{(X_2-X)^2+(Y_2-Y)^2} \quad (4)$$

$$\varepsilon_\alpha = \varepsilon_\phi - \varepsilon_\theta$$

The estimate values of the navigation variables $(\rho, \alpha, \phi)$ are:

$$\hat{\rho} = \sqrt{(X_2-\hat{X})^2+(Y_2-\hat{Y})^2}$$

$$\hat{\phi} = \operatorname{atan2}(Y_2-\hat{Y}, X_2-\hat{X}) - \theta_2 \quad (5)$$

$$\hat{\alpha} = \operatorname{atan2}(Y_2-\hat{Y}, X_2-\hat{X}) - \hat{\theta}$$

The input disturbances of the translational and angular velocities are defined by $\varepsilon_v$, $\varepsilon_\omega$; $\varepsilon_v \leq \|\varepsilon_v\|, \varepsilon_\omega \leq \|\varepsilon_\omega\|$, where $\|\varepsilon_v\|, \|\varepsilon_\omega\|$ are the absolute maximum of the input disturbances. With the existing of input disturbances, (2) is rewritten as follows:

$$\begin{cases} \dot{\rho} = -(v+\varepsilon_v)\cos\alpha \\ \dot{\alpha} = -(\omega+\varepsilon_\omega) + (v+\varepsilon_v)\dfrac{\sin\alpha}{\rho} \\ \dot{\phi} = (v+\varepsilon_v)\dfrac{\sin\alpha}{\rho} \end{cases} \quad (6)$$

Without loss of generality, we assume that the goal desired configuration of the system is $(X_d, Y_d, \theta_d) = (0,0,0)$ which can also be expressed by $(\rho_d, \alpha_d, \phi_d) = (0,0,0)$. The goal of the paper is to establish a stabilization control law for the mobile robot that is robust against input disturbances and measurement noises.

## 3. SYSTEM STATE ESTIMATION

Our approach for the proposed problem is the development of a closed-loop controller in which the feedback state are estimated by using an extended Kalman filter (EKF). The control law is then derived by constructing appropriate Lyapunov functions with constraints that asymptotically stabilize the system. Fig.2 shows the diagram of the controller.

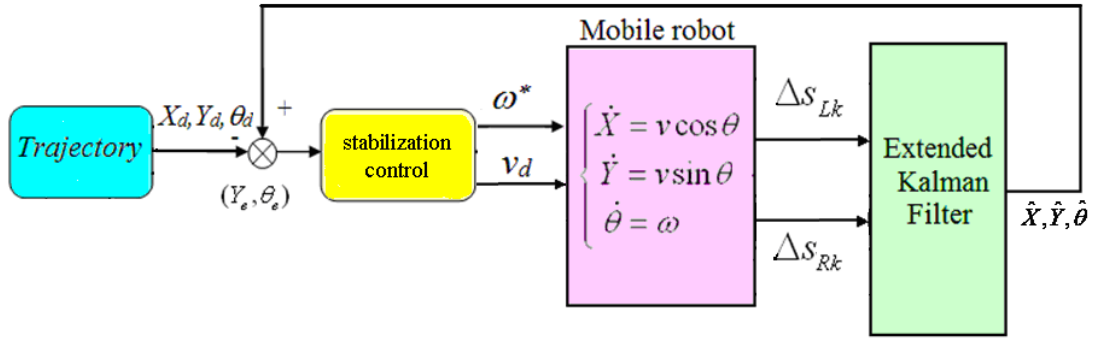

*Figure 2*. The control model

The design of the stabilization control block will be discussed in the next section. In this section, details of the state estimation using EKF is presented.

During one sampling period $\Delta t$, the rotational speed of the left and right wheels $\omega_L$ and $\omega_R$ create corresponding increment distances $\Delta s_L$ and $\Delta s_R$ traveled by the left and right wheels of the robot respectively:

$$\Delta s_L = \Delta t R \omega_L \qquad \Delta s_R = \Delta t R \omega_R \tag{7}$$

These can be translated to the linear incremental displacement of the robot's center $\Delta s$ and the robot's orientation angle $\Delta \theta$:

$$\Delta s = \frac{\Delta s_L + \Delta s_R}{2} \qquad \Delta \theta = \frac{\Delta s_R - \Delta s_L}{L} \tag{8}$$

The coordinates of the robot at time $k+1$ in the global coordinate frame can be then updated by:

$$\begin{bmatrix} x_{k+1} \\ y_{k+1} \\ \theta_{k+1} \end{bmatrix} = \begin{bmatrix} x_k \\ y_k \\ \theta_k \end{bmatrix} + \begin{bmatrix} \Delta s_k \cos(\theta_k + \Delta \theta_k / 2) \\ \Delta s_k \sin(\theta_k + \Delta \theta_k / 2) \\ \Delta \theta_k \end{bmatrix} \tag{9}$$

In practice, (9) is subjected to unavoidable disturbances which can be both systematic such as the imperfectness of robot model and nonsystematic such as the slip of wheels. These errors have accumulative characteristic so that they can break the stability of the system. EKF is an optimal solution for the problem.

Let $\mathbf{x} = [x\, y\, \theta]^T$ be the state vector. This state can be observed by some absolute measurements, $\mathbf{z}$. These measurements are described by a nonlinear function, $h$, of the robot coordinates and an independent Gaussian noise process, $\mathbf{v}$. Denoting the function (9) is $f$, with an input vector $\mathbf{u}$, the robot can be described by:

$$\begin{aligned} \mathbf{x}_{k+1} &= f(\mathbf{x}_k, \mathbf{u}_k, \mathbf{w}_k) \\ \mathbf{z}_k &= h(\mathbf{x}_k, \mathbf{v}_k) \end{aligned} \tag{10}$$

where the random variables $\mathbf{w}_k$ and $\mathbf{v}_k$ represent the process and measurement noise respectively. They are assumed to be independent to each other, white, and with normal probability distributions: $\mathbf{w}_k \sim \mathbf{N}(0, \mathbf{Q}_k) \quad \mathbf{v}_k \sim \mathbf{N}(0, \mathbf{R}_k) \quad E(\mathbf{w}_i \mathbf{v}_j^T) = 0$

The steps to calculate the EKF are then realized as below:

1. Prediction step with time update equations:

$$\hat{\mathbf{x}}_k^- = f(\hat{\mathbf{x}}_{k-1}, \mathbf{u}_{k-1}, \mathbf{0}) \tag{11}$$

$$\mathbf{P}_k^- = \mathbf{A}_k \mathbf{P}_{k-1} \mathbf{A}_k^T + \mathbf{W}_k \mathbf{Q}_{k-1} \mathbf{W}_k^T \tag{12}$$

where $\hat{\mathbf{x}}_k^- \in \Re^n$ is the *priori* state estimate at step $k$ given knowledge of the process prior to step $k$, $\hat{\mathbf{P}}_k^-$ denotes the covariance matrix of the state-prediction error, $\mathbf{A}_k$ is the Jacobian matrix of partial derivates of $f$ to $x$:

$$\mathbf{A}_{ij} = \frac{\partial \mathbf{f}_i}{\partial \hat{\mathbf{x}}_{pj(k-1)}}\bigg|_{(\hat{\mathbf{x}}_{p(k-1)}, \mathbf{u}_{(k-1)})} \Rightarrow \mathbf{A}_k = \begin{bmatrix} 1 & 0 & -\Delta s_k \sin(\theta_k + \Delta\theta_k/2) \\ 0 & 1 & \Delta s_k \cos(\theta_k + \Delta\theta_k/2) \\ 0 & 0 & 1 \end{bmatrix} \quad (13)$$

**W** is the Jacobian matrix of partial derivates of *f* to *w*:

$$\mathbf{W}_{ij} = \frac{\partial \mathbf{f}_i}{\partial \mathbf{w}_{j(k-1)}}\bigg|_{(\hat{\mathbf{x}}_{p(k-1)}, \mathbf{u}_{(k-1)})} \Rightarrow \mathbf{W}_k = \begin{bmatrix} -\Delta S_k \sin(\theta_k + \Delta\theta_k/2)/2l & \Delta S_k \sin(\theta_k + \Delta\theta_k/2)/2l \\ \Delta S_k \cos(\theta_k + \Delta\theta_k/2)/2l & -\Delta S_k \cos(\theta_k + \Delta\theta_k/2)/2l \\ 1/l & -1/l \end{bmatrix} \quad (14)$$

and $\mathbf{Q}_{k-1}$ is the input-noise covariance matrix which depends on the standard deviations of noise of the right-wheel rotational speed and the left-wheel rotational speed. They are modeled as being proportional to the rotational speed $\omega_{R,k}$ and $\omega_{L,k}$ of these wheels at step *k*. This results in the variances equal to $\delta\omega_R^2$ and $\delta\omega_L^2$, where $\delta$ is a constant determined by experiments. The input-noise covariance matrix $\mathbf{Q}_k$ is defined as:

$$\mathbf{Q}_k = \begin{bmatrix} \delta\omega_{R,k}^2 & 0 \\ 0 & \delta\omega_{L,k}^2 \end{bmatrix} \quad (15)$$

2. Correction step with measurement update equations:

$$\mathbf{K}_k = \mathbf{P}_k^- \mathbf{H}_k^T (\mathbf{H}_k \mathbf{P}_k^- \mathbf{H}_k^T + \mathbf{R}_k)^{-1} \quad (16)$$

$$\hat{\mathbf{x}}_k = \hat{\mathbf{x}}_k^- + \mathbf{K}_k \left( \mathbf{z}_k - \mathbf{h}(\hat{\mathbf{x}}_k^-) \right) \quad (17)$$

$$\mathbf{P}_k = (\mathbf{I} - \mathbf{K}_k \mathbf{H}_k) \mathbf{P}_k^- \quad (18)$$

where $\hat{\mathbf{x}}_k \in \Re^n$ is the *posteriori* state estimate at step *k* given measurement $\mathbf{z}_k$, $\mathbf{K}_k$ is the Kalman gain, **H** is the Jacobian matrix of partial derivates of *h* to *x*, $\mathbf{R}_k$ is the covariance matrix of measurement noise.

From (16), (17), (18), the estimated state is better than the raw measurement data and the variation in the estimation is reduced in each step to reach a stationary state. This estimation is the input for equations (6) and is essential for the controller design.

## 4. CONTROLLER DESIGN

Let $\Omega = \{(X, Y, \theta) : \rho, \alpha, \phi \in R\}$ be the set of all accessible configurations of the robot in the configuration space. Let $\Omega_l = \{(X, Y, \theta) : \rho(X, Y) < \varepsilon_P \cap |\phi(X, Y) - \alpha(X, Y, \theta)| < \varepsilon_\theta\}$ be defined as the local configuration set of the robot close to the goal configuration. Let $\Omega_g = \Omega - \Omega_l$ be the global configuration set of the robot distant from the goal configuration. In this section, we derive the control law for the motion control in the global and local configurations.

### 4.1. Stable control in the global configuration

Let the Lyapunov function for the global configuration set be given by

$$V_g = V_{g1} + V_{g2} = \frac{\rho^2}{2} + \frac{(\alpha^2 + h\phi^2)}{2} > 0 \tag{19}$$

From (19), $V_g$ is always positive. If we find the constraint of the inputs $v(t)$ and $\omega(t)$ so that $\dot{V}_g = \dot{V}_{g1} + \dot{V}_{g2}$ is always negative then the system is asymptotically stable and the control law successfully drives the robot to the destination.

Let $v = k_v \rho \cos\alpha$, the term $\dot{V}_{g1}$ becomes

$$\begin{aligned}\dot{V}_{g1} = \rho\dot{\rho} &= -k_v \rho^2 \cos^2\alpha + k_v \rho^2 \varepsilon_\alpha \cos\alpha \sin\alpha - \rho\varepsilon_v \cos\alpha \\ &\leq -k_v \rho^2 \cos^2\alpha + k_v \rho^2 \|\varepsilon_\alpha\|.|\cos\alpha \sin\alpha| + \rho\|\varepsilon_v\|.|\cos\alpha|\end{aligned} \tag{20}$$

We can choose a sufficiently large gain $K_v$ so that $k_v \rho^2 \cos^2\alpha$ is much more dominant than the terms $k_v \rho^2 \|\varepsilon_\alpha\|.|\cos\alpha \sin\alpha|$ and $\rho\|\varepsilon_v\|.|\cos\alpha|$. (20) becomes $\dot{V}_{g1} \leq 0$ in the region of $\Omega_g$ which implies that the term $V_{g1}$ converges to a nonnegative finite limit. Consider the term $\dot{V}_{g2}$:

$$\dot{V}_{g2} = \alpha\left[-\omega - \varepsilon_\omega + \left(\frac{k_v \rho \cos\alpha \sin\alpha}{\alpha} + \frac{k_v \rho \varepsilon_\alpha \cos^2\alpha}{\alpha} + \frac{\varepsilon_v \sin\alpha}{\alpha}\right)\frac{(\alpha + h\phi)}{\rho + \varepsilon_\rho}\right] \tag{21}$$

Let $\omega = k_\alpha \alpha + \dfrac{k_v \rho \cos\alpha \sin\alpha}{\alpha}\dfrac{(\alpha + h\phi)}{\rho}$, the term $\dot{V}_{g2}$ becomes:

$$\dot{V}_{g2} = -k_\alpha \alpha^2 - \alpha\varepsilon_\omega - \frac{k_v(\alpha + h\phi)\varepsilon_\rho}{2(\rho + \varepsilon_\rho)}\sin 2\alpha + \frac{k_v \rho(\alpha + h\phi)\varepsilon_\alpha}{\rho + \varepsilon_\rho}\cos^2\alpha + \frac{\varepsilon_v(\alpha + h\phi)\sin\alpha}{\rho + \varepsilon_\rho} \tag{22}$$

The term $k_\alpha \alpha^2$ is much more dominant than the terms $|\alpha\varepsilon_\omega|$, $\left|\dfrac{k_v(\alpha + h\phi)\varepsilon_\rho}{2(\rho + \varepsilon_\rho)}\sin 2\alpha\right|$, $\left|\dfrac{k_v \rho(\alpha + h\phi)\varepsilon_\alpha}{\rho + \varepsilon_\rho}\cos^2\alpha\right|$, $\left|\dfrac{\varepsilon_v(\alpha + h\phi)\sin\alpha}{\rho + \varepsilon_\rho}\right|$ so $\dot{V}_{g2} \leq 0$.

We have shown that when the robot is in the configuration set $\Omega_g$, the derivative of the Lyapunov function $\dot{V}_g \leq 0$ becomes semi-definite negative. As a result, by using the control law (23), the robot, which initially starts from the global configuration set $\Omega_g$, will be rendered to the local configuration set $\Omega_l$. The control law in the global configuration set $\Omega_g$ is rewritten as follows:

$$v = k_v \rho \cos\alpha \qquad \omega = k_\alpha \alpha + \frac{k_v \rho \cos\alpha \sin\alpha}{\alpha}\frac{(\alpha + h\phi)}{\rho} \tag{23}$$

**4.2. Stable control in the local configuration**

The control law (23) is asymptotically stable in the global configuration $\Omega_g$. It however is not stable in the local configuration $\Omega_l$. This can be proven as follows.

Assume that the navigation variable $\rho$ goes to small parameters $\varepsilon_p > \|\varepsilon_v\|/k_v$. The variables $(\alpha, \phi)$ go to their small disturbances ($\varepsilon_\alpha$, $\varepsilon_\phi$). The system kinematics (6) becomes:

$$\begin{cases} \dot{\rho} = -k_v \varepsilon_P + \varepsilon_v \\ \dot{\alpha} = \left[ -k_\alpha - \frac{k_v h \phi}{\alpha} - \frac{\varepsilon_\omega}{\alpha} + \frac{-k_v \varepsilon_\rho + \varepsilon_v}{(\varepsilon_P + \varepsilon_\rho)} + \frac{(k_v \varepsilon_P + \varepsilon_v)\varepsilon_\alpha}{(\varepsilon_P + \varepsilon_\rho)\alpha} \right] \alpha \\ \dot{\phi} = (k_v \varepsilon_P + \varepsilon_v) \frac{\alpha + \varepsilon_\alpha}{(\varepsilon_P + \varepsilon_\rho)} \end{cases} \quad (24)$$

Because $\varepsilon_P > \|\varepsilon_v\|/k_v$, we get

$$\dot{V}_{g1} = -k_v \rho^2 \cos^2 \alpha + k_v \rho^2 \varepsilon_\alpha \cos\alpha \sin\alpha - \rho \varepsilon_v \cos\alpha \leq -k_v \varepsilon_P^2 + \varepsilon_P \|\varepsilon_v\| \leq 0 \quad (25)$$

In (25), $V_{g1}$ is bounded; thus, $\rho$ is also bounded. However, when $\left( -k_\alpha - \frac{k_v h \phi}{\alpha} - \frac{\varepsilon_\omega}{\alpha} + \frac{-k_v \varepsilon_\rho + \varepsilon_v}{(\varepsilon_P + \varepsilon_\rho)} + \frac{(k_v \varepsilon_P + \varepsilon_v)\varepsilon_\alpha}{(\varepsilon_P + \varepsilon_\rho)\alpha} \right) > 0$ then $\alpha$ diverges from zero causing the system to be unstable. In the rest of the paper, we will re-design the control law to obtain the robustness property of the closed loop system.

For the local configuration set $\Omega_l$, let a Lyapunov function be given as:

$$V_l = \frac{\rho^2}{2} + \frac{(\theta - \theta_d)^2}{2} > 0 \quad (26)$$

Let $\theta_e = \theta - \theta_d$ and $\dot{\theta}_e = \dot{\theta} - \dot{\theta}_d = \omega$, the derivative of $V_l$ becomes:

$$\dot{V}_l = \rho \dot{\rho} + \theta_e \dot{\theta}_e = v\rho \cos\alpha + \theta_e \omega \quad (27)$$

Let the control law of the for the local configuration set $\Omega_l$ be given as follows:

$$v = -k_v \rho \cos\alpha \qquad \omega = -k_\theta \theta_e \quad (28)$$

where $k_v, k_\theta$ is positive gains. The term $\dot{V}_l$ becomes:

$$\dot{V}_l = -k_v \rho^2 \cos^2\alpha - k_\theta \theta_e^2 \leq 0 \quad (29)$$

This implies that the configuration of the robot will not escape to the larger values of ($\varepsilon_p$, $\varepsilon_\alpha$, $\varepsilon_\phi$) when the robot configuration is in the set $\Omega_l$ and the system is again stable.

## 5. SIMULATION AND EXPERIMENTS

To evaluate the functioning operation of the EKF-based state estimation and the stabilization controller, several simulations and experiments have been conducted.

### 5.1. Simulation results

Simulations are carried out in MATLAB in which the parameters are extracted from the real system[26]. The behaviors of the control law derived from the Lyapunov function in both configuration sets $\Omega_g$ and $\Omega_l$ are investigated. In the simulations, the initial configuration of the robot is $(0,0,0)$ and the goal configurations are $(2,2,30^0)$, $(2,2,60^0)$ and $(2,2,90^0)$ respectively. The absolute maximum values of measurement noises and input disturbances are as follows: $\varepsilon_p = \varepsilon_\phi = \varepsilon_\alpha = 0.001$; $\varepsilon_v = \varepsilon_\omega = 0.001$; the parameters for the controller are set as follows: $k_v = 10$, $k_\phi = 100$. Fig.3 shows the simulation results in which the final configurations of the robot are converged to the position $(2,2)$ from three different directions. This implies the success of the controller.

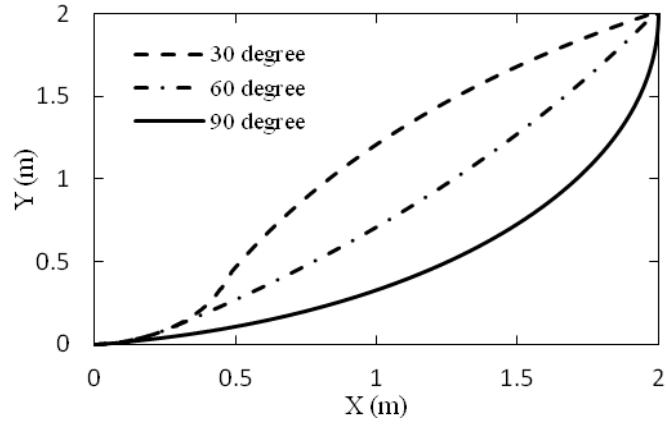

*Figure 3.* Stabilization results with the assumption of random measurement noises and input disturbances

### 5.2. Experimental results

*A) Experimental Setup*

Experiments on a real mobile robot are implemented in a rectangular shaped flat-wall environment constructed from several wooden plates surrounded by a cement wall is setup. The robot is a two wheeled, differential-drive mobile robot. Its wheel diameter is *10 cm* and the distance between two drive wheels is *60cm*. The drive motors are controlled by microprocessor-based electronic circuits. Due to the critical requirement of accurate speed control, the PID algorithm is implemented. The stability of motor speed checked by a measuring program written by LABVIEW is *±5%*. In case of straight moving, the speed of both wheels is set to *0.3m/s*. In turning, the speed of one wheel is reduced to *0.05m/s* in order to force the robot to turn to that wheel direction. The sensors employed as measurements include a compass sensor and a laser range finder. The compass sensor has the accuracy of *0.1⁰*. The LRF has the accuracy of *30mm* in distance and *0.25⁰* in deflect angle. The sampling time *ΔT* of the EKF is *100ms*.

*B) State estimate evaluation*

In order to evaluate the efficiency of the estimation algorithm, different configurations of the EKF were implemented [27]. Fig.4 describes the trajectories of a robot movement estimated by four methods: the odometry, the EKF with compass sensor, the EKF with LRF, and the EKF with the combination of LRF and compass sensor. The deviations between each trajectory and the real one are represented in fig.5.

In another experiment, the robot is programmed to follow predefined paths under two different scenarios: with and without the EKF. Fig.6a represents the trajectories of the robot moving along a rounded rectangular path in which the one with dots corresponds to the non-existence of EKF and the one with pluses corresponds to the existence of EKF in the implementation [26]. The trajectories in case the robot follows a rounded triangular path is shown in fig.6b. It is concluded that the EKF algorithm improves the robot localization and its combination with LRF and compass sensor gives the optimal result.

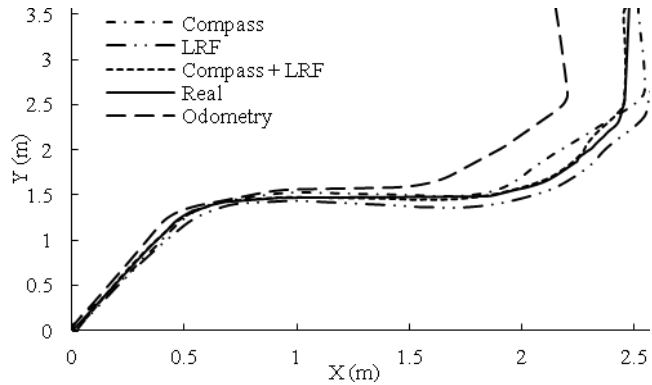

*Figure 4.* Estimated robot trajectories under different EKF configurations [27]

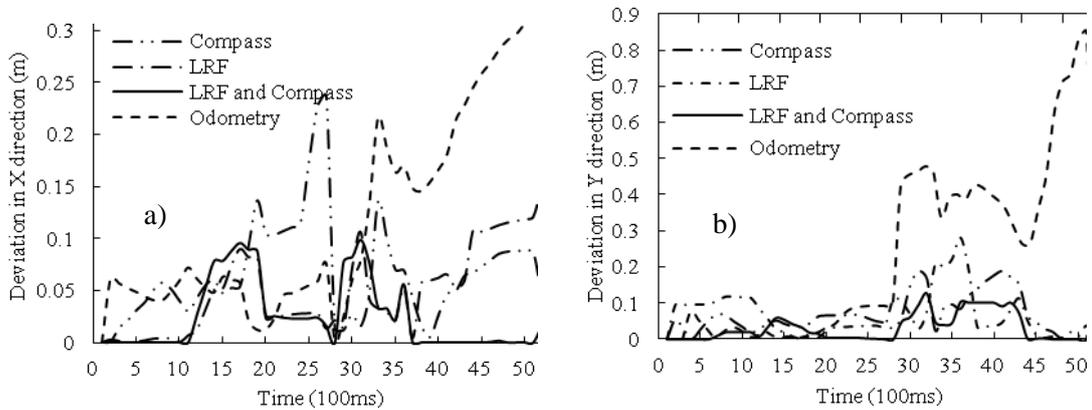

*Figure 5.* The deviation in X and Y direction between estimated positions and the real one
a) Deviation in X direction   b) Deviation in Y direction

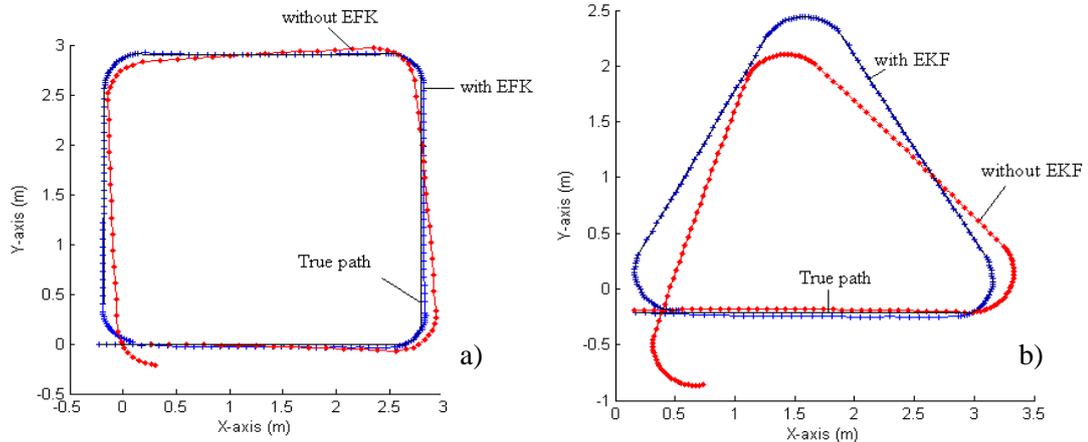

*Figure 6.* Trajectories of the robot moving along predefined paths with and without EKF

    a) A rounded rectangular path [26]    b) A rounded triangular path

*C) Stablization control*

In this experiment, we evaluate the applicability of the proposed controller in a real autonomous navigation application. The goal is to navigate the mobile robot from the starting point (0,0,0) to respectively reach the following destinations: $(2,2,30^0)$, $(2,2,60^0)$, $(2,2,90^0)$. Fig.8a describes the trajectories of the robot. Fig.7b and fig.7c are respectively the tangent velocity and direction variation of the robot during the operation. They are shown that the robot successfully reach the destinations and at those positions, the tangent velocity and direction variation go toward zero.

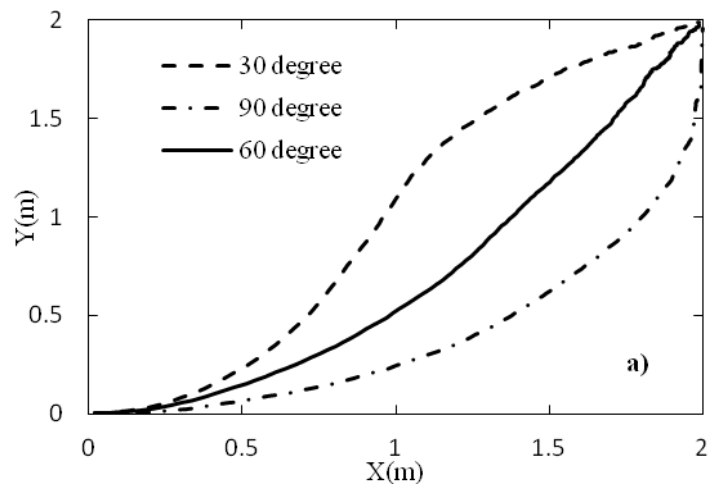

A

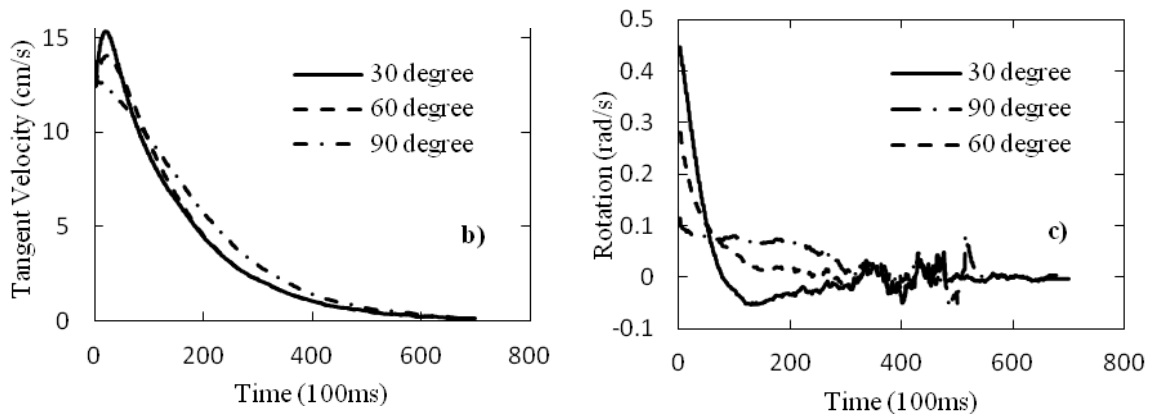

*Figure 7.* Results of stabilization control:

a) Trajectories of the robot   b) Tangent velocity   c) Direction variation

## 6. CONCLUSION

In this paper, the stabilization control problem of a mobile robot is formulated in the presence of system noise and measurement disturbances. A state estimation algorithm using EKF is proposed in which the knowledge of the system dynamics and the measurement information are combined in an optimal manner. Two Lyapunov functions corresponding to each subset configuration of the mobile robot is constructed and the control law is derived. The asymptotical stabilization of the system is theoretically analyzed and proven. Simulations and experiments confirm the validity of the proposed approach.